\documentclass[sigconf]{acmart}

\renewcommand\footnotetextcopyrightpermission[1]{}

\usepackage{graphicx}
\usepackage{amsmath}
\usepackage{amsthm}
\usepackage{booktabs}
\usepackage{algorithm}
\usepackage{algorithmic}
\usepackage{threeparttable}

\usepackage{enumitem}

\AtBeginDocument{%
  }
    
\begin{document}

\title{Human Mobility Prediction with Causal and Spatial-constrained Multi-task Network}

\author{Zongyuan Huang}
\email{herozen@sjtu.edu.cn}
\affiliation{
    \institution{Shanghai Jiao Tong University}
    \city{Shanghai}
    \country{China}
}

\author{Shengyuan Xu}
\email{xushengyuan@sjtu.edu.cn}
\affiliation{
    \institution{Shanghai Jiao Tong University}
    \city{Shanghai}
    \country{China}
}

\author{Menghan Wang}
\email{menghawang@ebay.com}
\affiliation{
    \institution{eBay Inc.}
    \city{Shanghai}
    \country{China}
}

\author{Hansi Wu}
\email{hanswu@ebay.com}
\affiliation{
    \institution{eBay Inc.}
    \city{Shanghai}
    \country{China}
}

\author{Yanyan Xu}
\authornote{Corresponding authors.}
\email{yanyanxu@sjtu.edu.cn}
\affiliation{
    \institution{Shanghai Jiao Tong University}
    \city{Shanghai}
    \country{China}
}

\author{Yaohui Jin}
\authornotemark[1]
\email{jinyh@sjtu.edu.cn}
\affiliation{
    \institution{Shanghai Jiao Tong University}
    \city{Shanghai}
    \country{China}
}

\renewcommand{\shortauthors}{Zongyuan Huang et al.}

\begin{abstract}
   Modeling human mobility helps to understand how people are accessing resources and physically contacting with each other in cities, and thus contributes to various applications such as urban planning, epidemic control, and location-based advertisement. Next location prediction is one decisive task in individual human mobility modeling and is usually viewed as sequence modeling, solved with Markov or RNN-based methods. However, the existing models paid little attention to the logic of individual travel decisions and the reproducibility of the collective behavior of population. To this end, we propose a Causal and Spatial-constrained Long and Short-term Learner (CSLSL) for next location prediction. CSLSL utilizes a causal structure based on multi-task learning to explicitly model the "\textit{when$\rightarrow$what$\rightarrow$where}", a.k.a. "\textit{time$\rightarrow$activity$\rightarrow$location}" decision logic. We next propose a spatial-constrained loss function as an auxiliary task, to ensure the consistency between the predicted and actual spatial distribution of travelers' destinations. Moreover, CSLSL adopts modules named Long and Short-term Capturer (LSC) to learn the transition regularities across different time spans. Extensive experiments on three real-world datasets show promising performance improvements of CSLSL over baselines and confirm the effectiveness of introducing the causality and consistency constraints. 
   The implementation is available at https://github.com/urbanmobility/CSLSL.
\end{abstract}


\maketitle

\section{Introduction}
\label{sec:intro}
    Human mobility modeling aims to explore the regularities and patterns of human behavior~\cite{gonzalez2008understanding,barbosa2018human} and plays a significant role in numerous applications, such as urban planning~\cite{xu2021emergence}, travel demand management~\cite{ccolak2016understanding,xu2018planning}, health risk assessment~\cite{xu2019unraveling}, epidemic spreading modeling and control~\cite{arenas2020modeling,jia2020population}, and so on. In the big data era, the accessibility to GPS, mobile phone records, and location-based social networks (LBSNs) provides an unprecedented chance to understand and model human mobility~\cite{barbosa2018human,luca2020deep}.
    
    In the research community of human mobility, physicists focus on statistical analysis from a macroscopic perspective and have summarized empirical rules~\cite{barbosa2018human}. For example, they found that, truncating the power law distribution can well fit the displacement distribution~\cite{gonzalez2008understanding}; despite the significant differences in the travel patterns, a majority of users' mobility behaviors are predictable~\cite{song2010limits}.
    Computer scientists, on the other hand, prefer to model the transition regularities from location sequences, using Markov models~\cite{rendle2010factorizing}, recurrent neural networks (RNNs)~\cite{feng2018deepmove}, etc. In summary, statistical physics study the collective behavior at population level, while deep learning methods emphasize modeling individual travel trajectories. Thus, we can expect that integrating physical domain knowledge into a deep learning model encourages the model to pay attention to group behaviors and promotes the performance of deep learning models at population level.
    
    Here we place our emphasis on next location prediction, a vital task in human mobility modeling at individual level. A body of work leverages machine learning methods to tackle this problem due to the sequential nature of mobility behavior. A common thread of these studies is to efficiently capture behavior patterns from sparse data~\cite{luca2020deep,lian2020geography,guo2020attentional,luo2021stan}. Traditional methods mainly adopt Markov chains to model transition probability matrices across locations, along with techniques like factorization~\cite{rendle2010factorizing,cheng2013you, he2016inferring} or metric embedding~\cite{feng2015personalized}. 
    In recent years, deep learning methods are gaining increasing attention in next location prediction as the recurrent neural network (RNN) presents its capability to capture sequential dependency. 
    To model multi-scale spatio-temporal periodicity, researchers designed attention or gate mechanisms and introduced time and distance interval information~\cite{feng2018deepmove,manotumruksa2018contextual,zhao2019go,sun2020go,yang2020location}.
    Also a few studies incorporate semantic information such as location categories to cope with the data sparsity~\cite{he2017category,yu2020category,guo2020attentional}. 
    However, methods that capture dependencies only from location sequences are difficult to fully fit complex human travel behaviors, especially with sparse data.

\begin{figure}[bt]
        \begin{center}
        \includegraphics[width=0.4\textwidth]{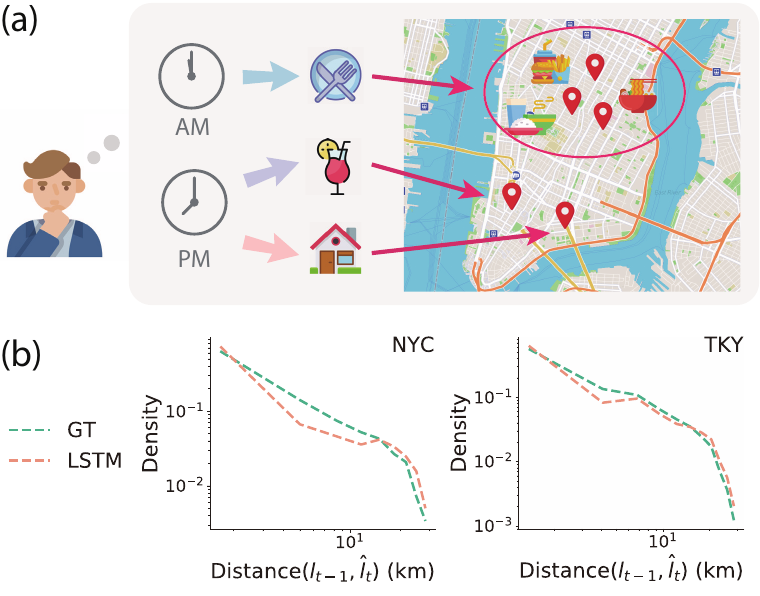}
        \end{center}
        \caption{Illustrations for causality and spatial consistency in human mobility. (a) An example to explain the "\textit{when$\rightarrow$what$\rightarrow$where}" decision logic. (b) The deviation of the modeled displacement distribution via LSTM from the ground truth (GT) in New York City and Tokyo.}
        \label{fig:motivation}
    \end{figure}

    To tackle this challenge, we seek to integrate physical knowledge into deep learning methods to enhance the capability of human mobility prediction. Specifically, we propose two physical constraints.
    The first one is summarized as "\textit{when$\rightarrow$what$\rightarrow$where}" causal relationship. 
    "\textit{When}", "\textit{what}", and "\textit{where}" are the three core elements of human travel behavior and the dependencies between them can explain the motivation of location transfer. For example, as shown in Figure~\ref{fig:motivation}(a), people have specific demands at different times, causing the shifts between locations. 
    Considering causal dependencies enables more comprehensive modeling of human mobility. 
    The second constraint is the macro-statistical characteristics reflecting group behavior. Figure~\ref{fig:motivation}(b) illustrates the deviation of the modeled displacement distribution via LSTM from the true distribution in New York City and Tokyo, suggesting LSTM is more likely to focus on shorter trips with higher frequency. 
    Ensuring the consistency between the model output and the macro-statistical characteristics is expected to improve the model's capability to fit travel behavior.
    We summarize these two constraints as causality and consistency constraints and incorporate them into deep learning models.
    
    To this end, we propose a Causal and Spatial-constrained Long and Short-term Learner (CSLSL), a model integrating the decision logic and the consistency constraints of human mobility modeling.
    To model the "\textit{when$\rightarrow$what$\rightarrow$where}" decision logic, we introduce a causal structure in CSLSL. Based on a multi-task learning, the causal structure utilizes three similar network branches to model the regularities of time, activity, and location, respectively. 
    In line with the "\textit{when$\rightarrow$what$\rightarrow$where}" logic, we explicitly build connections between the three branches in the causal structure.
    As for consistency, we exploratively propose a spatial-constrained loss to reduce the distance between the predicted and actual locations, and indirectly ensure the consistency of the spatial density distribution. In addition, we adopt a Long and Short-term Capturer (LSC) to learn the transition patterns of different time spans. There are two units in LSC that focus on long-term and short-term regularities respectively. 
    
    The main contributions of this work are summarized as follows:
    \begin{itemize}[leftmargin=*]
    \item We propose CSLSL to integrate the travel decision logic and the macro-statistical consistency for human mobility modeling. To our best knowledge, CSLSL is the first model to learn the causality and consistency constraints for next location prediction. 
    \item We introduce a causal structure that can capture not only the separate regularities of time, activity, and location, but also the "\textit{when$\rightarrow$what$\rightarrow$where}" causal dependencies. In this way, CSLSL models more essential travel logic in addition to sequence relationships.
    \item To ensure the consistency in spatial distribution, we propose a spatial-constrained loss to reduce the gap between the predicted and actual destinations.
    \item We evaluate CSLSL on three real-world datasets to confirm the performance improvements. We also conduct ablation studies and visualization analyses of results such as displacement distribution to demonstrate the effectiveness of our design. 
    \end{itemize}

\section{Related Work}
\label{sec:related}
    
\subsection{Next Location Prediction}
    Here we classify the approaches to the next location prediction problem into two categories: traditional and deep learning methods.
    Traditional methods mainly apply Markov chain (MC) and focus on constructing a better location transition probability matrix~\cite{rendle2010factorizing,cheng2013you,feng2015personalized,zhao2016stellar}. 
    For instance, factorized personalized Markov chain (FPMC) combines the matrix factorization technique with Markov chains to learn users' personalized transition matrices~\cite{rendle2010factorizing}. The limitation of the MC-based methods lies in the difficulty in capturing long-term and high-order regularity~\cite{guo2020attentional,luo2021stan}.
    
    Deep learning methods have advantages of learning dense representation and complex dependency. Recently, RNN-based methods show promising performance in mining sequential information. A popular scheme of deep learning methods is incorporating time and distance intervals to assist the model in learning the spatio-temporal regularities of human mobility. 
    Specifically, these methods integrate spatio-temporal information into hidden state transition~\cite{liu2016predicting}, gate mechanisms~\cite{kong2018hst,manotumruksa2018contextual,zhao2019go}, or self-attention mechanisms~\cite{lian2020geography,luo2021stan}, and exploit spatio-temporal contexts in a implicit manner. 
    To leverage the spatio-temporal contexts, researchers explicitly used spatial and temporal factors as attention weights to select the historical hidden states~\cite{yang2020location,zhao2020discovering}. 
    Another scheme emphasizes the long-term patterns of human behavior, such as DeepMove~\cite{feng2018deepmove} and LSTPM~\cite{sun2020go}. They introduce two different components to model long-term and short-term preferences respectively. 
    There is also another scheme that utilizes semantic information such as location categories to improve the performance of location prediction~\cite{yu2020category,guo2020attentional,wang2021spatio}.
    However, methods that focus on modeling location transfer patterns in sequences cannot effectively capture complex human decision logic. 
    In our work, we propose a causal structure to explicitly capture the "\textit{when$\rightarrow$what$\rightarrow$where}" decision logic.

\subsection{Time- or Activity-jointed Location Prediction}
    The methods that jointly predict time or activity learn knowledge from related tasks to improve the prediction performance of location.
    RMTPP~\cite{du2016recurrent} combines RNN and temporal point process (TPP) to jointly model the time and location information. He et al.~\cite{he2017category} proposed a two-fold approach that predicts category with Bayesian Personalized Ranking (BPR) technique and then predicts the category-based location. Krishna et al.~\cite{krishna2018lstm} utilized two distinct LSTM networks to predict activities and durations. DeepJMT~\cite{chen2020context} fuses spatio-temporal information and social context to predict time and location with a hierarchical RNN and TPP technique. Sun et al.~\cite{sun2021joint} proposed a hybrid LSTM and a sequential LSTM with a self-attention mechanism to jointly model location and travel time.  
    The limitation of these approaches is that they attempt to implicitly and passively learn the correlation between time, category, and location information, but this relationship is explicit and can be directly exploited. In contrast, CSLSL explicitly models the causal dependencies between time, category, and location information through two structural designs.

\subsection{Statistical Physics-informed human mobility modeling}   

    Explicitly integrating knowledge of statistical physics  contributes to guiding model optimization and improving the performance of machine learning methods. On the task of trajectory generation, researchers introduced knowledge of statistical physics to constrain the macroscopic performance of their models, such as the individual trajectory generation model TimeGeo~\cite{jiang2016timegeo} and DITRAS~\cite{pappalardo2018data}, and flow generation model DeepGravity~\cite{simini2021deep}. 
	Unlike the trajectory generation task, only a limited amount of work on individual mobility prediction incorporates knowledge of statistical physics. 
    Zhao et al.~\cite{zhao2020discovering} integrated domain knowledge, specifically a power-law decay for distances and an exponential decay for time intervals, into an attention mechanism to adjust the impact of historical information on current prediction. However, deep learning-based methods focus more on fitting individual behavior while neglecting group behavior constraints described by macro-statistical characteristics, for example, the regional attractiveness of city blocks. The spatial distribution of model prediction results should be consistent with the actual statistical distribution. Predicted locations closer to the actual location are more expected. Toward this, we propose a spatial-constrained loss function to narrow the distance between the predicted and actual locations, thereby ensuring the consistency of the spatial distribution.

\section{Problem Formulation}
   A person's travel behavior can be represented as a sequence of locations, associated with the timestamps and her user ID. In LBSN datasets, each location is also associated with its functional category to support the analysis of user's activity. Let $\mathcal{U} = \{u_1, \dots, u_{|\mathcal{U}|}\}$, $\mathcal{L} = \{l_1, \dots, l_{|\mathcal{L}|}\}$ and $\mathcal{C} = \{c_1, \dots, c_{|\mathcal{C}|}\}$ denote a set of users, locations and functional categories, respectively. Each location $l_i$ is associated with its category and geographical coordinate $(c_i, lat_i, lon_i)$.
   
   \textbf{Definition 1 (Record).}  Record $r$ is a 3-tuple $(u_i, l_j, t_k)$, representing that the user $u_i$ visited location $l_j$ at time $t_k$, where $u_i\in\mathcal{U}, l_j\in\mathcal{L}$.
   
   \textbf{Definition 2 (Individual Trajectory)}. A person's trajectory is defined as a record sequence $\mathcal{R}=\{r_1, r_2, \dots, r_{|\mathcal{R}|}\}$, which consists of the person's all records arranged in chronological order. Note that the time interval between two consecutive records is heterogeneous due to the irregular travel behavior.   
   
   \textbf{Definition 3 (Session)}. Session $S$ is a subsequence of records in a time slot. One user's trajectory $\mathcal{R}$ can be split into a series of sessions with various strategies. 
   For example, DeepMove adopts a specific time interval between two consecutive records to split the trajectory~\cite{feng2018deepmove}. 
   Other strategies segment users' trajectories using a fixed number of records~\cite{yang2020location,luo2021stan} or a meaningful time window such as days, weeks, etc~\cite{sun2020go}.
   We define the session where the prediction target is located as the short-term session $S_p$ and the previous historical sessions as long-term sessions $\{S_q\},q\in\{1,\dots,p-1\}$. 
   
   The location prediction problem is formulated as: given a record sequence of a user $\mathcal{R}_{t-1}=\{r_1, \dots, r_{t-1}\}$, the goal is to predict where the user $u$ is most likely to go in her next trip.  
   We use $\hat{l}_t$ to denote the predicted next location. Note that the timestamp of the next trip $t$ is also unknown.
   
\section{Methodology}

\begin{figure}[bt]
    	\setlength{\abovecaptionskip}{3mm}
        \begin{center}
        \includegraphics[width=0.4\textwidth]{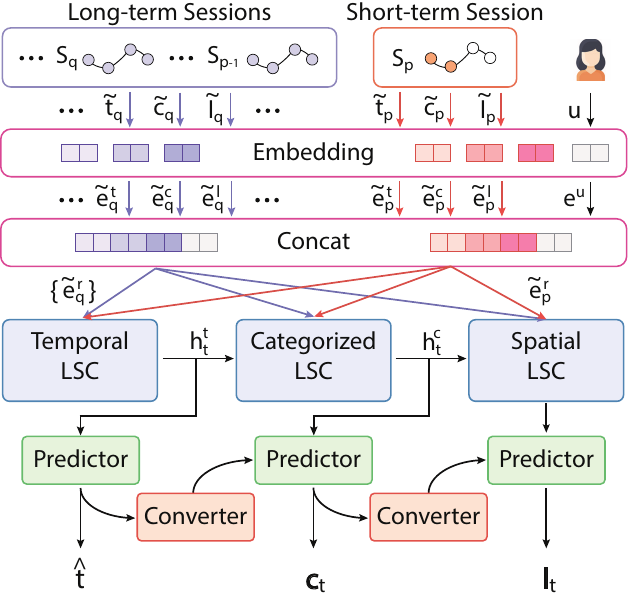}
        \end{center}
        \caption{The architecture of the proposed CSLSL model.}
        \label{fig:framework}
        \vspace{-3mm}
    \end{figure}

    In this section, we first analyze the causality and consistency constraints in human mobility modeling, and then elaborately introduce the design of the proposed model, Casual and Spatial-constrained Long and Short-term Learner (CSLSL). The architecture of CSLSL is illustrated in Figure~\ref{fig:framework}. It mainly consists of two parts, an embedding part for learning the representations of arrival time, category, and location, from users' recent and historical records; and the second part for learning the regularities of mobility behavior in a multi-task learning based causal module and making predictions.
    
\subsection{Causality and Consistency Constraints}
\label{sec:41}
    A common practice for next location prediction is to discover similar subsequences or location transition relationships from historical records. This is accomplished by integrating the context information such as distance or time intervals into attention or RNN-centered framework~\cite{zhao2019go,yang2020location,sun2020go,luo2021stan}.
    We can formulate such mainstream scheme as $P(\hat{l}_t|\mathcal{R}_{t-1})$, where $\mathcal{R}_{t-1}$ is the historical record sequence.
    Another scheme adopts multi-task learning techniques to jointly predict next location with time or activity~\cite{chen2020context,sun2021joint}, formulated as $P(\hat{l}_t, \hat{c}_t, \hat{t}|\mathcal{R}_{t-1})=P(\hat{l}_t|\mathcal{R}_{t-1})P(\hat{c}_t|\mathcal{R}_{t-1})P(\hat{t}|\mathcal{R}_{t-1})$, where we assume that the location category can approximate the type of activity.
    Although these two schemes combine contextual information to capture hidden regularities of location transition, they ignore the causal dependencies in the context information. 
    
    As aforementioned, we regard "\textit{when}", "\textit{what}" and "\textit{where}" as three crucial elements to describe human mobility~\cite{zhang2021putting,pacheco2022predictability}. "\textit{When}" refers to the time the trip takes place, e.g. "midday". "\textit{What}" tells about the activities people participate in and also answers the reasons for the trip, such as "having lunch". "\textit{Where}" is the destination of the trip, like "steakhouse". Periodic activities exist in human mobility and occur at specific times and places, such as going to work in the morning and going to a restaurant for lunch, which reveals the correlation between the three elements. 
    When we mention a specific timestamp, we have various activity choices. But we are accustomed to doing certain activities at certain times, such as going to the gym in the evening. Similarly, one activity (category) corresponds to multiple locations (POIs), while one location ID only corresponds to one activity, also reflected in the dataset. Moreover, our target is location prediction, thus location should be the final subtask to leverage the predicted time and activity information. 
    Therefore, we summarize a "\textit{when$\rightarrow$what$\rightarrow$where}", a.k.a. "\textit{time$\rightarrow$activity$\rightarrow$location}" causal relationship, which is in line with the coarse-to-fine logic of the human decision. 
    The proposed scheme can be formulated as:
    \begin{equation}
        P(\hat{l}_t, \hat{c}_t, \hat{t}|\mathcal{R}_{t-1})=P(\hat{l}_t|\hat{c}_t, \hat{t}, \mathcal{R}_{t-1})P(\hat{c}_t|\hat{t}, \mathcal{R}_{t-1})P(\hat{t}|\mathcal{R}_{t-1}).
    \end{equation}
    The scheme explicitly models the dependencies between time, activity, and location, and alleviates the difficulty of location prediction.
    For example, people are accustomed to going to restaurants at midday instead of bar, that is, $P(\hat{c}_t=restaurant|\hat{t}=midday,\mathcal{R}_{t-1})>P(\hat{c}_t=bar|\hat{t}=midday,\mathcal{R}_{t-1})$. Each individual has her personalized $P(\hat{l}_t|\hat{c}_t=restaurant, \hat{t}=midday, \mathcal{R}_{t-1})$, and the casual constrained location distribution is easier to learn than $P(\hat{l}_t|\mathcal{R}_{t-1})$. In CSLSL, we introduce a causal structure to implement the scheme. In experiments, we also demonstrate that "time->acitivity->location" outperforms "activity->time->location" and further discuss the reason.
    
    On the other side, integrating physical knowledge provides more information and prior constraints to guide the optimization of deep learning models~\cite{karniadakis2021physics,willard2020integrating}. In human mobility modeling, one can expect that properly introducing the physical laws and domain knowledge would narrow down the gap between the output of deep learning-based approaches and the observed macro-statistical characteristics of human behavior. 
    Due to the difficulty in applying statistical constraints in the training of deep learning models, here we consider the geographic spatial consistency in an indirect way. Specifically, we devise a loss function to constrain the distance between the predicted and actual locations. That is, the closer the predicted location is to the ground truth, the smaller loss we have. By this way, we can indirectly ensure the consistency of the displacement distribution and the consistency of the spatial distribution of travelers' destinations.

\subsection{Long and Short-term Capturer}
    Human travel behavior has long and short-cycle repetitive patterns, such as going to work every day and going to the supermarket once a week. Inspired by DeepMove~\cite{feng2018deepmove} and LSTPM~\cite{sun2020go}, we devise a Long and Short-term Capturer (LSC) to learn the behavioral patterns in different observation cycles. In the whole framework shown in Figure~\ref{fig:framework}, we apply three LSCs to model the time, activity, and location sequences, respectively.
    
    Let $\boldsymbol{e}^l\in\mathbb{R}^{d^l},\boldsymbol{e}^c\in\mathbb{R}^{d^c},\boldsymbol{e}^t\in\mathbb{R}^{d^t}$ and $\boldsymbol{e}^u\in\mathbb{R}^{d^u}$ denote the embedded representation of location, category, time and user, respectively. Given a historical record sequence $\mathcal{R}$, CSLSL embeds each record as $(\boldsymbol{e}^l, \boldsymbol{e}^c, \boldsymbol{e}^t, \boldsymbol{e}^u)$ in hidden spaces. Note that we first convert the continuous timestamp as the hour in a day $t^h$ and the day in a week $t^d$ to present the daily and weekly periodicity. By doing so, we have $\boldsymbol{e}^t=\boldsymbol{e}^{t^h}\oplus \boldsymbol{e}^{t^d}$. Then these representations in a record are concatenated together, $\boldsymbol{e}^r=\boldsymbol{e}^l\oplus \boldsymbol{e}^c\oplus \boldsymbol{e}^t\oplus \boldsymbol{e}^u$.
    We next split each user's record sequence into multiple sessions with a certain time window, like days or weeks. The records in short-term session and long-term sessions are represented as $\widetilde{\boldsymbol{e}}^r_p=\{\boldsymbol{e}^r_1, \dots,\boldsymbol{e}^r_{t-1}\}$ and $\{\widetilde{\boldsymbol{e}}^r_q\}=\{\widetilde{\boldsymbol{e}}^r_1, \dots,\widetilde{\boldsymbol{e}}^r_{p-1}\}$, respectively. 
    
    \begin{figure}[bt]
        \begin{center}
        \includegraphics[width=0.38\textwidth]{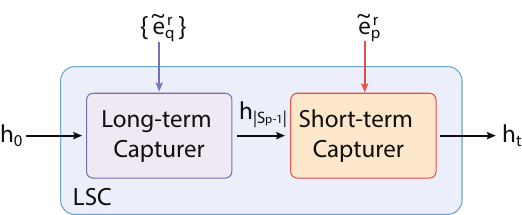}
        \end{center}
        \caption{The illustration of the LSC module.}
        \label{fig:module}
        \vspace{-3mm}
    \end{figure}

    Our proposed LSC consists of two capturers to learn the transition regularities in the short-term and long-term sessions, respectively, as shown in Figure~\ref{fig:module}. We formulate LSC as:
    \begin{equation}
    \label{eq:lsc-1}
        \boldsymbol{h}_t = LSC(\widetilde{\boldsymbol{e}}^r_p, \{\widetilde{\boldsymbol{e}}^r_q\}, \boldsymbol{h}_{0}),
    \end{equation}
    where $\boldsymbol{h}_{0}$ is the initial hidden state. In the LSC structure, the short-term capturer takes the hidden state $h_{|S_{p-1}|}$ of the long-term capturer as the initial hidden state to combine the historical information. Because GRU is simple but efficient in modeling temporal data, we apply a layer of GRU in both of the capturers:
    \begin{equation}
    \label{eq:gru-1}
        \boldsymbol{h}_i = GRU(\boldsymbol{e}^r_{i-1}, \boldsymbol{h}_{i-1}),
    \end{equation}
    where $i\in\{1,2,\dots,|S_{p-1}|\}$ for long-term capturer and $i\in\{1,2,\dots,t\}$ for short-term capturer.

\subsection{Causal Structure}

    To model the "\textit{time$\to$activity$\rightarrow$location}" logic relationship discussed in subsection~\ref{sec:41}, we introduce a causal structure based on multi-task learning techniques. As illustrated in Figure~\ref{fig:framework}, we utilize three branches with the same architecture to model the change patterns of time, activity, and location, respectively. 
    Specifically, in each branch, we convey the same record representations to the LSC module and then transfer the output hidden states to the predictor. To explicitly model the summarized causal relation in human travel behavior, we next design two paths for information transfer between various tasks. The first path lies between two LSC modules, passing on the task-specific hidden states. The second path lies between two predictors. In this path, the predicted result of the upstream task is processed by the converter and then conveyed to the downstream task. Here we use the fully connected layer as the predictor (P) and converter(C). That is $\boldsymbol{y} = Linear(\boldsymbol{x})=\boldsymbol{W}\boldsymbol{x}+\boldsymbol{b}$.

    Mathematically, the branch of "\textit{time}" is formulated as:
    \begin{equation}
        \boldsymbol{h}^t_t = LSC(\widetilde{\boldsymbol{e}}^r_p, \{\widetilde{\boldsymbol{e}}^r_q\}, 0),
    \end{equation}
    \begin{equation}
        \hat{t} = Linear^{(P^t)}(\boldsymbol{h}^t_t),
    \end{equation}
    where $\boldsymbol{W}^{(P^t)}\in\mathbb{R}^{1\times|\boldsymbol{h}^t|}$, $\boldsymbol{h}^t_t$ is the hidden state of the next time, and $\hat{t}$ is the predicted time.
    As the downstream task of "\textit{time}" in causal structure, the branch of "\textit{activity}" can be formulated as:
    \begin{equation}
        \boldsymbol{h}^c_t = LSC(\widetilde{\boldsymbol{e}}^r_p, \{\widetilde{\boldsymbol{e}}^r_q\}, \boldsymbol{h}^t_t),
    \end{equation}
    \begin{equation}
        \boldsymbol{c}_t = Linear^{(P^c)}(\boldsymbol{h}^c_t \oplus Linear^{(C^t)}(\hat{t})),
    \end{equation}
    \begin{equation}
        \hat{c}_t = argmax(\boldsymbol{c}_t),
    \end{equation}
    where $ \boldsymbol{W}^{(C^t)}\in\mathbb{R}^{|\boldsymbol{e}^t|\times1}, \boldsymbol{W}^{(P^c)}\in\mathbb{R}^{|\mathcal{C}|\times(|\boldsymbol{h}^c|+|\boldsymbol{e}^t|)}$, $\boldsymbol{h}^c_t$ is the hidden state of the next activity, and $\hat{c}_t$ is the predicted activity.
    Eventually, we can formulate the branch of "\textit{location}" as:
    \begin{equation}
        \boldsymbol{h}^l_t = LSC(\widetilde{\boldsymbol{e}}^r_p, \{\widetilde{\boldsymbol{e}}^r_q\}, \boldsymbol{h}^c_t),
    \end{equation}
    \begin{equation}
        \boldsymbol{l}_t = Linear^{(P^l)}(\boldsymbol{h}^l_t \oplus Linear^{(C^c)}(\boldsymbol{c}_t)),
    \end{equation}
    \begin{equation}
        \hat{l}_t = argmax(\boldsymbol{l}_t),
    \end{equation}
    where $ \boldsymbol{W}^{(C^c)}\in\mathbb{R}^{|\boldsymbol{e}^c|\times|\mathcal{C}|}, \boldsymbol{W}^{(P^l)}\in\mathbb{R}^{|\mathcal{L}|\times(|\boldsymbol{h}^l|+|\boldsymbol{e}^c|)}$, $\boldsymbol{h}^l_t$ is the hidden state of the next location, $\boldsymbol{l}_t$ is the distribution of predicted location, and $\hat{l}_t$ is the predicted location.

\begin{figure*}[bt]
    \begin{center}
    \includegraphics[width=0.9\textwidth]{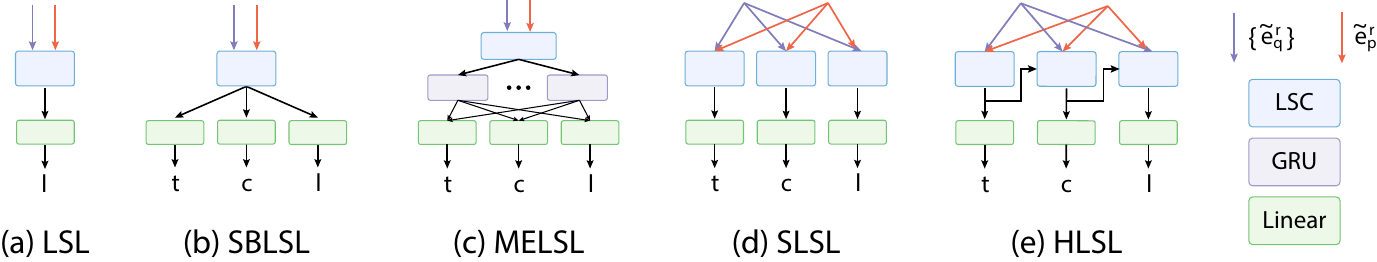}
    \end{center}
    \caption{Structure variation. (a) Long and Short-term Learner (LSL); (b) Share-Bottom LSL (SBLSL); (c) Multi-Experts LSL (MELSL); (d) Separate LSL (SLSL); (e) Hierarchical LSL (HLSL). Note the gate networks of MELSL are not drawn for simplicity.}
    \label{fig:variation}
\end{figure*}

\subsection{Spatial-constrained Loss Function}

    As discussed in subsection~\ref{sec:41}, for seeking an alignment of spatial distribution of destinations, we propose a spatial-constrained loss function to shorten the distance from the predicted location to the ground truth at individual level. The distance constraint can be regarded as a self-supervised auxiliary task, integrating the geographical information and restricting the candidate set for better next location prediction. 
    The existing methods introduce the distance constraints in a regression subtask, directly predicting the geographical locations of destinations~\cite{xue2021mobtcast}. 
    However, in the classification scheme, we must query the coordinates of location IDs to calculate their distance. This operation is not derivable. 
    We get inspiration from REINFORCE~\cite{williams1992simple}, which introduces the reward in the loss function to train a policy network, and also consider the distance error as a coefficient to weight the cross-entropy between ground truth and the predicted location ID with their physical distance. 
    The spatial-constrained loss function is defined as:
    \begin{equation}
        L_s = - \sum^{N}_{i=1} distance(\hat{l}_{t,i}, l_{t,i}) \cdot log(\sigma(\boldsymbol{l}_{t,j})),
    \end{equation}
    where $N$ is the total number of records and $\sigma$ is the softmax function.

    We next employ MAE loss for time prediction and cross entropy loss for category and location prediction. Thus we have 
    $L_t=MAE(\hat{t},t)=\sum^{N}_{i=1}|\hat{t}_i-t_i|$ and 
    $L_*=CrossEntropy(*)=-\sum^{N}_{i=1} log(\sigma(\boldsymbol{*_{t,i}})),*\in\{c,l\}$. Thus, the total loss function can be written as
    \begin{equation}
        L_{total} = L_l + \lambda_t L_t + \lambda_c L_c + \lambda_s L_s,
    \end{equation}
    where $\lambda_t, \lambda_c, \lambda_s$ are weights for their loss functions. 

\subsection{Structure Comparison}
\label{sec:variant}

    There are various strategies for task combination in multi-task learning, such as share-bottom structure~\cite{ruder2017overview,qin2020multitask}, hierarchical structure~\cite{sanh2019hierarchical,chen2020context}, and multi-expert structure~\cite{ma2018modeling,qin2020multitask}. Inspired by these structures, we propose five variants, as shown in Figure~\ref{fig:variation}, to demonstrate the advantages of our causal structure. Note that these variants use the same basic components as CSLSL, such as GRUs and fully connected layers.
    
    Long and Short-term Learner (LSL) is a basic approach with only one branch to predict location. To jointly predict "\textit{time}", "\textit{activity}", and "\textit{location}", Share-Bottom LSL (SBLSL)
    introduces two additional predictors that share the same bottom LSC module with the original one. Multi-Experts LSL (MELSL) is an advanced version of SBLSL, with a similar structure to Mixture of Sequential Expert (MoSE)~\cite{qin2020multitask}. MELSL employs several GRUs as experts to focus on different aspects of sequence dependencies and gate networks to combine relevant aspects for each task.
    
    Unlike the share-bottom structure, Separate LSL (SLSL) employs a separate branch for each task and the only shared information between each task is the same record representations. 
    Considering the dependencies between tasks, Hierarchical LSL (HLSL) concatenates the record embedding and the output hidden state of the upstream task as the input of the downstream task. Thus the equation~\ref{eq:gru-1} changes to: 
    \begin{equation}
        \boldsymbol{h}^k_i = GRU(\boldsymbol{e}^r_{i-1} \oplus \boldsymbol{h}^{k-1}_i, \boldsymbol{h}^k_{i-1}),
    \end{equation}
    where $\boldsymbol{h}^k_i$ is the hidden state of the $k$-th task at $i$-th time step, and the equation~\ref{eq:lsc-1} changes to:
    \begin{equation}
        \boldsymbol{h}^k_t = LSC([\widetilde{\boldsymbol{e}}^r_p,\widetilde{\boldsymbol{h}}^{k-1}_p], \{[\widetilde{\boldsymbol{e}}^r_q,\widetilde{\boldsymbol{h}}^{k-1}_q]\}, 0),
    \end{equation}
    where $[\widetilde{\boldsymbol{e}}^r_p,\widetilde{\boldsymbol{h}}^{k-1}_p] = \{\boldsymbol{e}^r_1 \oplus \boldsymbol{h}^{k-1}_1 , \dots,\boldsymbol{e}^r_{t-1} \oplus \boldsymbol{h}^{k-1}_{t-1}\}$.

\section{Experiments}
\subsection{Data Description}
    We leverage three publicly available check-in datasets in the experiments: two datasets from Foursquare~\cite{yang2014modeling} in New York (NYC) and Tokyo (TKY) and one dataset from Gowalla~\cite{cho2011friendship} in Dallas. Data in NYC and TKY were collected from 3 April 2012 to 16 February 2013, and data in Dallas was collected from 4 October 2016 to 30 April 2017. The number of users, locations, records in three datasets are summarized in Table~\ref{tab:data}, where $|\mathcal{*}|$ denotes the number of $*$. The number of location categories $|\mathcal{C}|$ in NYC and TKY are 400 and 385, while Dallas does not contain category information.
    
    To prepare data for baselines and proposed models, we first filter out both users and locations with fewer than 10 records, in line with previous work~\cite{feng2018deepmove,sun2020go}. We then merge the consecutive records with the same user and location on the same day. The statistic information of the raw and processed data are depicted in Table~\ref{tab:data}. After pre-processing, the number of categories for NYC and TKY are reduced to 308 and 286. For CSLSL and its variations, we split trajectories into sessions according to week due to the data sparsity. In addition, we require that each session contains at least two records and a user contains at least five sessions to guarantee a training/testing split of 8/2, following~\cite{feng2018deepmove}.  
    All baselines have their own further data preparation strategies and the model-specific dataset information is also shown in Table~\ref{tab:data}. It's noteworthy that, LSTPM~\cite{sun2020go} requires at least three records in each session and Flashback~\cite{yang2020location} limits the minimum records of each user to 100. These practices filter out more sparse data and reduce the challenge of prediction. Moreover, GETNext requires category information as input, thus it cannot work on dataset Dallas.
    
    \begin{table}[hbt]
    \caption{Statistic information of raw and processed data.}       
    \label{tab:data} 
    \centering
    \scalebox{0.7}{
        \begin{tabular}{cccccccccccc}
            \hline 
                & \multicolumn{3}{c}{\textbf{NYC}}  
                & \multicolumn{3}{c}{\textbf{TKY}} 
                & \multicolumn{3}{c}{\textbf{Dallas}} \\
            \cline{2-4} \cline{5-7} \cline{8-10}
                & $|\mathcal{U}|$ & $|\mathcal{L}|$ & $|\mathcal{R}|$ 
                & $|\mathcal{U}|$ & $|\mathcal{L}|$ & $|\mathcal{R}|$ 
                & $|\mathcal{U}|$ & $|\mathcal{L}|$ & $|\mathcal{R}|$ \\
            \hline
                Raw & 1,083 & 38,333 & 227,428 & 2,293 & 61,858 & 573,703 & 5,894  & 5,767 & 167,016 \\
            \hline
                Processed & 1,083 & 4,638 & 139,183 & 2,293 & 7,222 & 427,746 & 2,412 & 5,642 & 146,117 \\
            \hline
                FPMC-D & 1,083 & 4,638 & 94,168 & 2,293 & 7,222 & 334,561 & 2,412 & 5,642 & 100,743 \\
                FPMC-W &  1,083 & 4,638 & 92,260 & 2,293 & 7,222 & 303,854  & 2,412 & 5,642 & 122,419 \\
                DeepMove & 1,061 & 4,627 & 111,968 & 2,284 & 7,206 & 333,215 & 1,193 & 5,346 & 93,911 \\
                Flashback & 439& 4,093 & 96,771 & 1,451 & 6,998 & 366,604 & 340   & 5,434 & 82,287\\
                LSTPM & 579 & 4,128 & 62,511 & 1,866 & 7,007 & 273,880 & 587 & 5,169 & 55,784\\
                GeoSAN & 1,073 & 4,611 & 138,229 & 2,289 & 7,209 & 427,157 & 2,300 & 5,357 & 142,980  \\
                STAN & 1,000 & 4,503 & 82,783 & 2,200 & 7,047 & 201,831 & 2,400 & 5,615 & 95,656  \\
                GETNext & 1,066 & 4,621 & 131,920 & 2,280 & 7,200 & 414,993 & - & - & -  \\
            \hline 
                Ours & 1,065 & 4,635 & 133,944 & 2,280 & 7,204 & 422,432 & 1,357 & 5,428 & 120,531 \\
            \hline
        \end{tabular}
    }
    \end{table}
    
\begin{table*}[hbt]
\caption{Performance comparison between baselines, CSLSL, its variants and ablations on three real-world datasets.}
\label{tab:performance}
    \centering
    \scalebox{0.92}{
    \begin{threeparttable}
    \begin{tabular}{ccccccc|cccccc|ccc}
        \hline
            &
            \multicolumn{6}{c|}{\textbf{NYC}} & \multicolumn{6}{c|}{\textbf{TKY}} & \multicolumn{3}{c}{\textbf{Dallas}} \\
        \cline{2-7} \cline{8-13} \cline{14-16}
            &
            \multicolumn{3}{c}{\textbf{Category}} & \multicolumn{3}{c|}{\textbf{Location}} &
            \multicolumn{3}{c}{\textbf{Category}} & \multicolumn{3}{c|}{\textbf{Location}} &
            \multicolumn{3}{c}{\textbf{Location}} \\
        \cline{2-4} \cline{5-7} \cline{8-10} \cline{11-13} \cline{14-16} 
            & 
            R@1 & R@5 & R@10 & R@1 & R@5 & R@10 & 
            R@1 & R@5 & R@10 & R@1 & R@5 & R@10 & 
            R@1 & R@5 & R@10   \\
        \hline
            FPMC-D & 0.217 & 0.510 & 0.631 & 0.173 & 0.424 & 0.526 & 
                    0.462 & 0.623 & 0.688 & 0.176 & 0.408 & 0.503 &
                    0.086 & 0.204 & 0.261 \\
            FPMC-W & 0.196 & 0.394 & 0.482 & 0.162 & 0.323 & 0.382 & 
                    0.428 & 0.513 & 0.557 & 0.117 & 0.245 & 0.313 &
                    0.055 & 0.125 & 0.161 \\
            DeepMove & 0.244 & 0.499 & 0.586 & 0.198 & 0.407 & 0.470 & 
                    0.398 & 0.567 & 0.626 & 0.161 & 0.331 & 0.400 & 
                    0.089 & 0.185 & 0.229 \\
            Flashback\tnote{*} & 0.280 & 0.618 & 0.737 & 0.223 & 0.521 & 0.639 &  
                    \underline{0.479} & 0.730 & 0.801 & 0.207 & \underline{0.486} & \textbf{0.583} & 
                    0.090 & 0.199 & 0.256 \\
            LSTPM\tnote{*} & \textbf{0.335} & \underline{0.659} & \textbf{0.765} & 0.267 & 0.560 & \textbf{0.662} &  
                    0.461 & 0.727 & 0.797 & 0.231 & 0.457 & 0.543 &   
                    \underline{0.123} & \textbf{0.247} & \textbf{0.316} \\
            GeoSAN & 0.193 & 0.433 & 0.602 &  0.166 & 0.430 & 0.584 &  
                    0.317 & 0.551 & 0.694 &  0.158 & 0.392 & 0.528 & 
                    0.078 & 0.186 & 0.265 \\
            STAN & 0.218 &  0.480 & 0.591  &  0.192  & 0.411 &  0.494 &  
                    0.376 & 0.575 & 0.668 &  0.167 & 0.388 & 0.478 & 
                    0.074 & 0.153 & 0.196 \\
            GETNext & 0.303 & 0.646 & 0.749 & 0.246 & 0.536 & 0.622 &  
                    0.452 & 0.758 & 0.844 & 0.216 & 0.456 & 0.550 & 
                    - & - & - \\
        \hline
            LSL  & 0.288 & 0.587 & 0.684 & 0.242 & 0.506 & 0.589 &  
                    0.446 & 0.697 & 0.762 & 0.225 & 0.462 & 0.548 &  
                    0.101 & 0.203 & 0.254 \\
            SBLSL & 0.290 & 0.595 & 0.682 & 0.242 & 0.488 & 0.571 &  
                    0.446 & 0.769 & 0.851 & 0.229 & 0.460 & 0.545 & 
                    0.107 & 0.194 & 0.234 \\
            MELSL & 0.275 & 0.586 & 0.662 & 0.227 & 0.432 & 0.526 &  
                    0.439 & 0.648 & 0.745 & 0.211 & 0.419 & 0.475 &
                    0.078 & 0.121 & 0.164 \\
            SLSL  & 0.281 & 0.611 & 0.722 & 0.253 & 0.534 & 0.632 &  
                    0.409 & 0.731 & 0.825 & 0.233 & 0.458 & 0.559 &
                    0.114 & 0.220 & 0.272 \\
            HLSL  & 0.296 & 0.628 & 0.734 & 0.256 & 0.536 & 0.625 &  
                    0.441 & 0.760 & 0.847 & 0.232 & 0.472 & 0.562 & 
                    0.115 & 0.230 & 0.284 \\
            CLSL-ctl & 0.315 & 0.647 & 0.745 & 0.257 & 0.543 & 0.636 &
            		  0.472 & 0.780 & 0.861 & 0.227 & 0.472 & 0.560 &
            		  - & - & - \\
        \hline
            CLSL & 0.322 & 0.658 & 0.747 & 0.261 & 0.553 & 0.643 &  
                   0.459 & \underline{0.782} & \underline{0.864} & \underline{0.238} & 0.476 & 0.567 & 
                   0.120 & 0.229 & 0.282 \\
            CSLSL-c & 0.315 & 0.638 & 0.757 & 0.247 & 0.546 & 0.643 &  
                    0.450 & 0.775 & 0.858 & 0.230 & 0.463 & 0.551 &  
                    - & - & - \\
            CSLSL-t & 0.319 & 0.648 & 0.749 & 0.264 & 0.556 & 0.652 &  
                    \underline{0.479} & 0.740 & 0.829 & 0.233 & 0.478 & 0.568 &
                    0.118 & 0.231 & 0.284 \\
            CSLSL & \underline{0.327} & \textbf{0.661} & \underline{0.759} & \textbf{0.268} & \textbf{0.568} & \underline{0.656} &  
                    \textbf{0.488} & \textbf{0.801} & \textbf{0.875} & \textbf{0.240} & \textbf{0.488} & \underline{0.580} & 
                    \textbf{0.126} & \underline{0.243} & \underline{0.297} \\
        \hline  
        \end{tabular}
        \begin{tablenotes}
        \footnotesize
        \item[*]Flashback and LSTPM filter out much more sparse users in their data preparation, reducing the challenge of prediction.  
        \end{tablenotes}
        \end{threeparttable}
    } 
\end{table*}

\subsection{Baselines and Settings}

    \textbf{Baselines.} We compare CSLSL with the state-of-the-art baselines: 
    \begin{itemize}[leftmargin=*, topsep=2pt]
        \item FPMC~\cite{rendle2010factorizing} is a Markov-based model that uses factorization to learn individual transition matrices.
        \item DeepMove~\cite{feng2018deepmove} adopts an attention mechanism to learn long-term preference and an GRU module to capture short-term preference. 
        \item Flashback~\cite{yang2020location} uses spatio-temporal distances as attention weights to search the historical hidden states for current prediction.
        \item LSTPM~\cite{sun2020go} considers temporal similarity and distance factor to model long-term preferences and geographical relevance to model short-term preferences. 
        \item GeoSAN~\cite{lian2020geography} designs a geography encoder to implicitly capture spatial proximity and introduces a loss function based on importance sampling to better use the informative negative samples. 
        \item STAN~\cite{luo2021stan} introduces a two-layer attention architecture with spatio-temporal relation matrices to explicitly capture the spatio-temporal correlations. 
        \item GETNext~\cite{yang2022getnext} utilizes a GCN to integrate collective movement patterns and a transformer encoder to capture transition regularities. Besides, it introduces location categories as inputs and prediction targets.
    \end{itemize}
    
    \textbf{Settings.} 
    For convincingly comparing these baselines with our CSLSL, we collected the open-source codes released by the authors and attempted to find the optimal hyper-parameters in the experiments.
    It's worth noting that most of the baselines only predict next location, without category and time of visitation. Thus, we match the predicted location ID to its category for comparison and exclude the performance comparison of time prediction. Besides, we split users' trajectories by day and week for FPMC model, referred to as FPMC-D and FPMC-W, respectively. 
    For CSLSL, the dimensions of representation vectors $\boldsymbol{e}^l,\boldsymbol{e}^c,\boldsymbol{e}^{t^h},\boldsymbol{e}^{t^d}$ and $\boldsymbol{e}^u$ are set to 200, 100, 10, 20, and 20 for all datasets. The dimension of the hidden state in all GRUs is set to 600. We use the Adam optimizer with the learning rate of 0.0001, and $\lambda_t$, $\lambda_c$, and $\lambda_s$ are set to $10$. 
    
    \textbf{Metrics.} In the next location prediction task, what we care about is whether the actual location is in the top $N$ of our predictions, $N=\{1,5,10\}$. 
    $Recall@N$ is the most commonly used metric and is equal to $Accuracy@N$ because we don't have false positive (FP) and true negative (TN).
    The definition of $Recall@N$ is 
    
    \begin{equation}
        Recall@N = \frac{1}{|\mathcal{U}|}\sum_{u\in\mathcal{U}}\frac{|\mathcal{L}_u^T\cap\mathcal{L}_u^P|}{|\mathcal{L}_u^T|},
    \end{equation}
    
    where $\mathcal{L}_u^T$ and $\mathcal{L}_u^P$ are the target and top $N$ prediction location sets, respectively.

\subsection{Performance Comparison with Baselines}

    The experimental results are averaged over 10 independent runs and shown in Table~\ref{tab:performance}. For each city, the results are presented in three pieces, representing the results of baselines (line 1-8), variants (line 9-14), and ablations (line 15-18), respectively. 
    The best performance in each column is highlighted in bold text and the second best one is underlined. For NYC and TKY, we present the predicted results for categories and locations, while for Dallas, we only show the location prediction results due to the lack of category information.
      
    From the experiment results, we can observe that the proposed CSLSL shows promising performances compared with baselines. In terms of $Recall@1$ in location prediction, CSLSL achieves $27\%$, $37\%$, and $43\%$ averaged performance improvements over these deep learning  baselines in three datasets. For $Recall@1$ in category prediction, the improvements are $34\%$ and $23\%$ in NYC and TKY, respectively. Considering $Recall@\{5,10\}$ in location prediction, CSLSL achieves similar performances with LSTPM and Flashback, which filter more than $46\%, 18\%,$ and $57\%$ of sparse users than we do on three datasets shown in Table~\ref{tab:data}. 
    These similar performances in the more challenging dataset settings can also reflect the superiority of our model.
    Disregarding these two models, CSLSL still obtains over $20\%$ averaged improvements than the rest of deep learning baselines. 
    Moreover, CSLSL has improved by 8.9\% and 11.1\% in Recall@1 in NYC and TKY compared to GetNext, which has similar dataset statistics to ours.
    The poor performances of all models on the Dallas dataset may be due to the data sparseness. Even so, CSLSL can still utilize the time-location relationship and the spatial constraints to achieve performance gains.
    
    Among the baselines, we observe that the overall performance of location prediction on TKY is lower than NYC. This is probably because the TKY dataset has a larger number of users and locations than NYC, increasing the difficulty of mobility predicting. However, CSLSL obtains more performance improvement on TKY than NYC for location prediction compared with baselines. For instance, the performance of CSLSL improves 8.9\% on NYC compared with GETNext, while the improvement is 11.1\% on TKY. On the other side, the category predictions for all models have higher accuracy on TKY than on NYC. We may conclude that the larger performance improvement on TKY than NYC mainly owes to the proper modeling of the dependencies.

\subsection{Performance Comparison with Variants}
\label{sec:res_variants}

    To fairly demonstrate the effectiveness of the proposed causal structure, here we develop an ablated version of CSLSL via dropping the spatial-constraint loss, namely CLSL, and compare it with the 5 variants discussed in the subsection~\ref{sec:variant}. 
    Moreover, we also consider the "\textit{what$\to$when$\to$where}" relationship, thus we change the order of these three branches in CLSL from "\textit{time$\to$category$\to$location}" to "\textit{category$\to$time$\to$location}" and this variant is named CLSL-ctl. 
    We present the results of the variants and CLSL in the second and third pieces of Table~\ref{tab:performance}. The category prediction results of LSL are obtained in the same way as the baselines. 
    
    Compared with LSL, SBLSL has a similar performance of location prediction and slightly improved performance of category prediction, suggesting that the shared bottom of SBLSL has indeed learned the category transfer regularities. However, these learned regularities make no contribution to the location prediction. 
    Besides, the performance of MELSL is weaker than LSL and SBLSL, which may be because MELSL does not clarify the relationship between tasks and its experts cannot find suitable optimization directions. 
    The better performance of SLSL than SBLSL indicates that the separate modules to learn transition relationships are better than the shared one. 
    HLSL achieves the best performance in the second pieces of Table~\ref{tab:performance}, suggesting that there are dependencies between \textit{time}, \textit{category}, and \textit{location}, and that capturing the dependencies facilitates location prediction.

    The performances of these variants are weaker than CLSL, suggesting that although these variants utilize temporal and categorical information, they cannot effectively and autonomously capture the dependencies between \textit{time}, \textit{category}, and \textit{location}. In contrast, the causal structure explicitly captures the dependencies between tasks in two ways, thereby fully exploiting their dependencies to improve performance. Moreover, the better performance of CLSL than CLSL-ctl is in line with expectations, because \textit{location} has stronger dependencies with \textit{category} than \textit{time} and category information can bring more performance gains for location prediction. 
    Therefore, our proposed causal structure explicitly models "\textit{time$\to$category$\to$location}" rather than "\textit{category$\to$time$\to$location}".

\subsection{Ablation Study}

    We also conduct ablation studies to examine the contributions of different components in CSLSL. The ablated models include:
    \begin{itemize}[leftmargin=*, topsep=2pt]
        \item LSL: the version that only keep the location branch.
        \item CLSL: the version that removes the spatial-constraint loss.
        \item CSLSL-t: the version that removes the time branch.
        \item CSLSL-c: the version that removes the category branch. 
    \end{itemize}
    
    As shown in Table~\ref{tab:performance}, we can find that CSLSL-t achieves better performance than CSLSL-c, indicating the "\textit{category$\to$location}" relationship has stronger dependency constraints than "\textit{time$\to$location}". This result is also consistent with what we discussed in the subsection~\ref{sec:res_variants}. 
    The best performance of the complete CSLSL demonstrates the significance of the entire "\textit{time$\to$category$\to$location}" decision logic. Comparing the performance of CLSL and CSLSL, we can confirm that the spatial-constraint loss function have a positive impact on the performance improvement. Moreover, LSL achieves decent performance compared with baselines, probably because it leverages category information and the LSC module is capable of capturing the long-term and short-term preferences.

\subsection{Results Visualization Analysis}

\begin{figure}[bt]
        \begin{center}
        \includegraphics[width=0.42\textwidth]{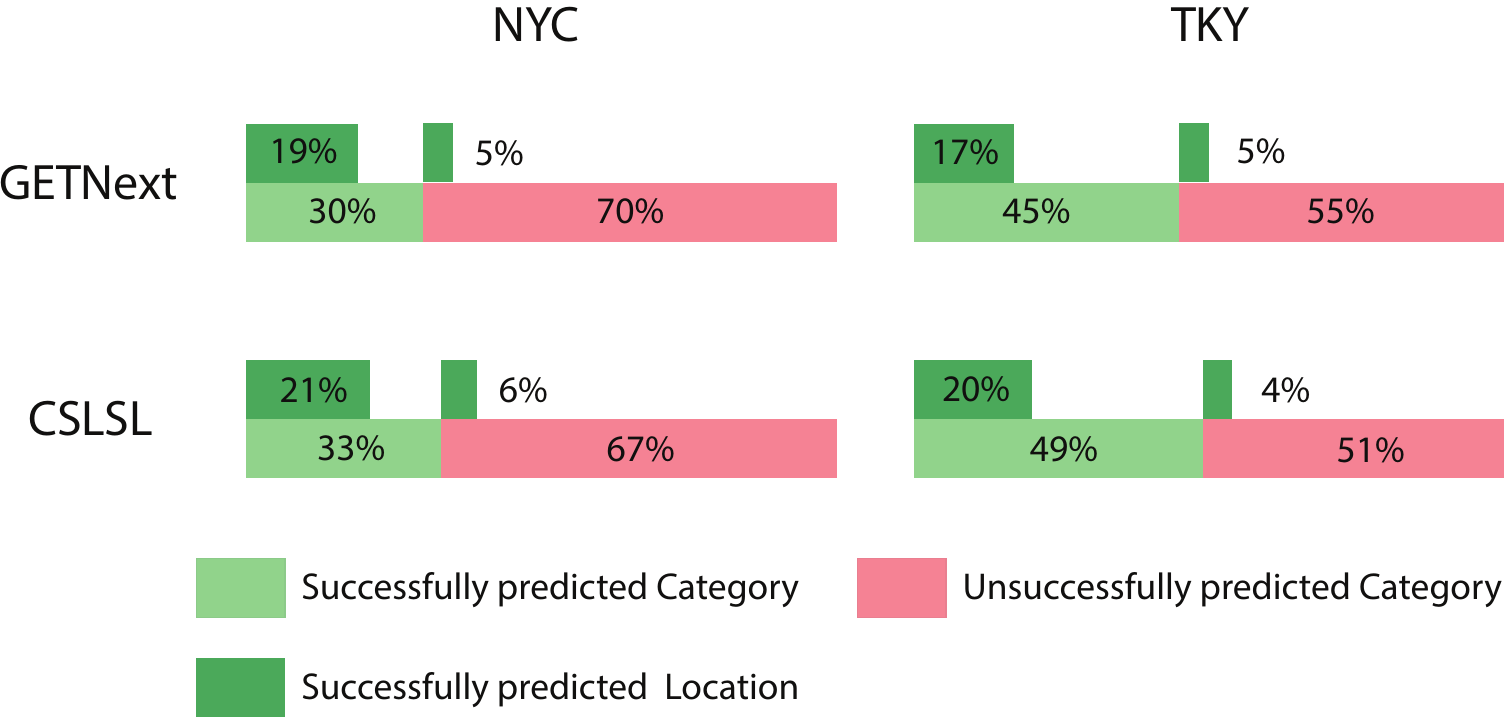}
        \end{center}
        \caption{Effect analysis of the causal structure.}
        \label{fig:CT}
    \end{figure}

    We conduct result visualization analysis to further understand the effectiveness of the causal structure and the spatial-constrained loss.
    For the causal structure, we compare the category and the location prediction results of GETNext and CSLSL, as shown in Figure~\ref{fig:CT}. We can observe that for CSLSL, the records with successfully predicted both categories and locations on NYC and TKY account for 21\% and 20\% of all records.
    Compared with GETNext, CSLSL successfully predicted 10\% and 18\% more locations with more accurately predicted categories on NYC and TKY, respectively.
    This intuitively indicates that the causal structure can enhance the location prediction with more accurate category prediction results. Interestingly, the location can be predicted correctly with an unsuccessfully predicted category. This is because the category information is introduced as additional auxiliary information without imposing mandatory constraints on the location prediction.

    \begin{figure}[bt]
        \begin{center}
        \includegraphics[width=0.46\textwidth]{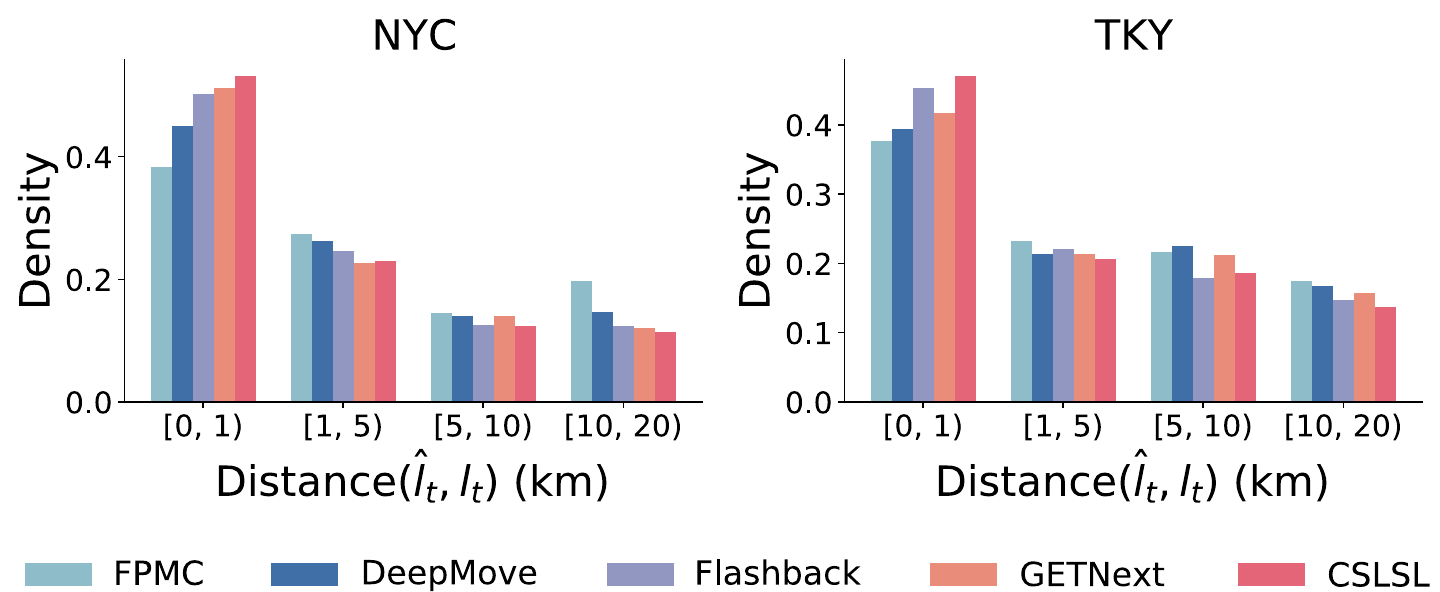}
        \end{center}
        \caption{Distance distribution between the predicted and target location.}
        \label{fig:distPT}
    \end{figure}

	\begin{figure}[bt]
        \begin{center}
        \includegraphics[width=0.46\textwidth]{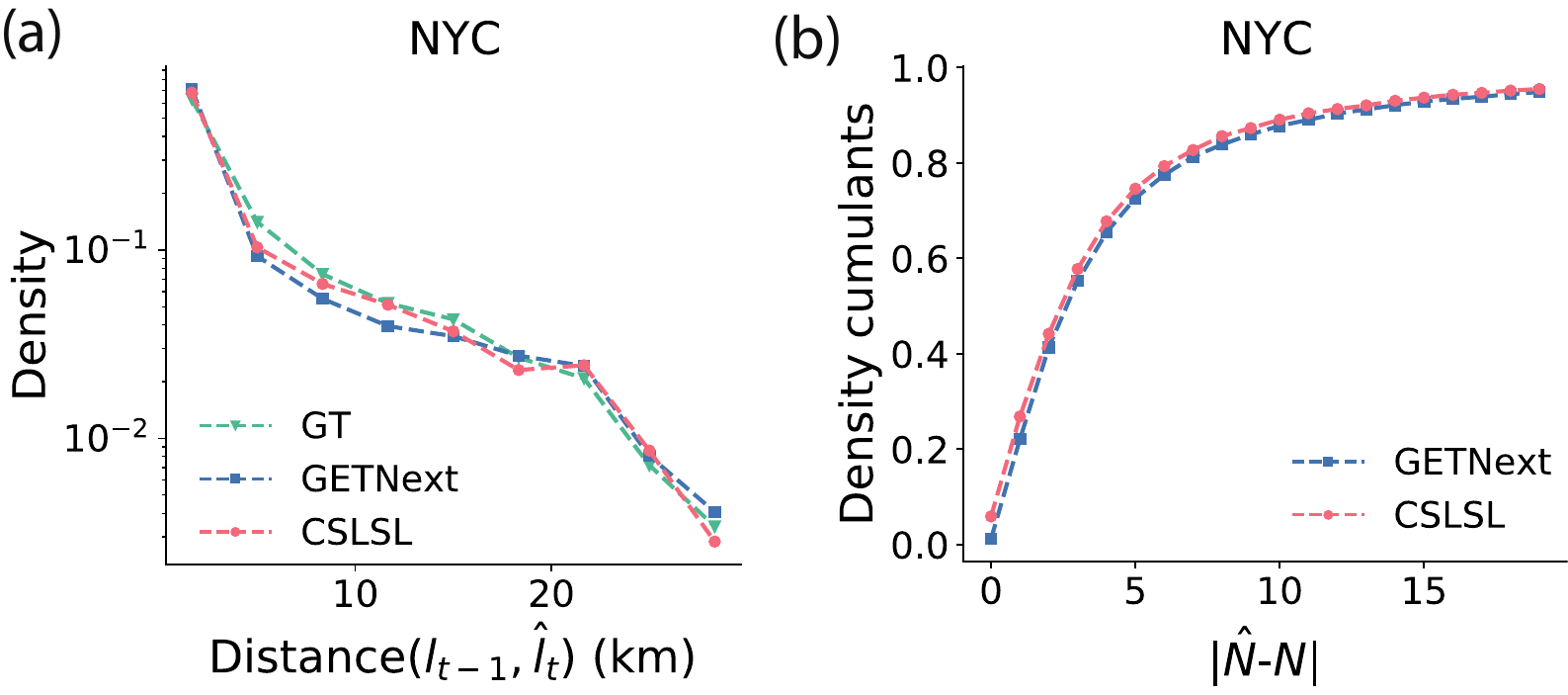}
        \end{center}
        \caption{(a) Comparison of the predicted displacement with the ground truth (GT). (b) Prediction error comparison of regional attractiveness.}
        \label{fig:ab}
    \end{figure}

    Regarding to the spatial-constrained loss, we examine whether the distances between predicted and actual locations are successfully constrained, and compare CSLSL with four baselines. As shown in Figure~\ref{fig:distPT}, the results show that the predicted locations of CSLSL is closer to the actual locations, which indicates that the proposed loss can successfully constrain the distance errors. 
    Furthermore, we inspect the constraining effect of the proposed loss on spatial consistency. Figure~\ref{fig:ab}(a) shows the comparison of the predicted displacement with the ground truth. It can be seen that the predicted displacement of CSLSL is closer to the true distribution. This is because the constraint between the predicted and actual locations can indirectly ensure the consistency of the predicted displacements and the ground truth. 
    Figure~\ref{fig:ab}(b) shows the prediction error of regional attractiveness. We divided the geographic regions into square grids with side length of 500m, and count the difference between the predicted and actual visits in each grid. As presented in Figure~\ref{fig:ab}(b), CSLSL has a smaller prediction error of regional attractiveness, suggesting that the proposed loss successfully constrains the spatial consistency.

\subsection{Sensitivity Analysis}

    \begin{figure}[bt]
        \begin{center}
        \includegraphics[width=0.48\textwidth]{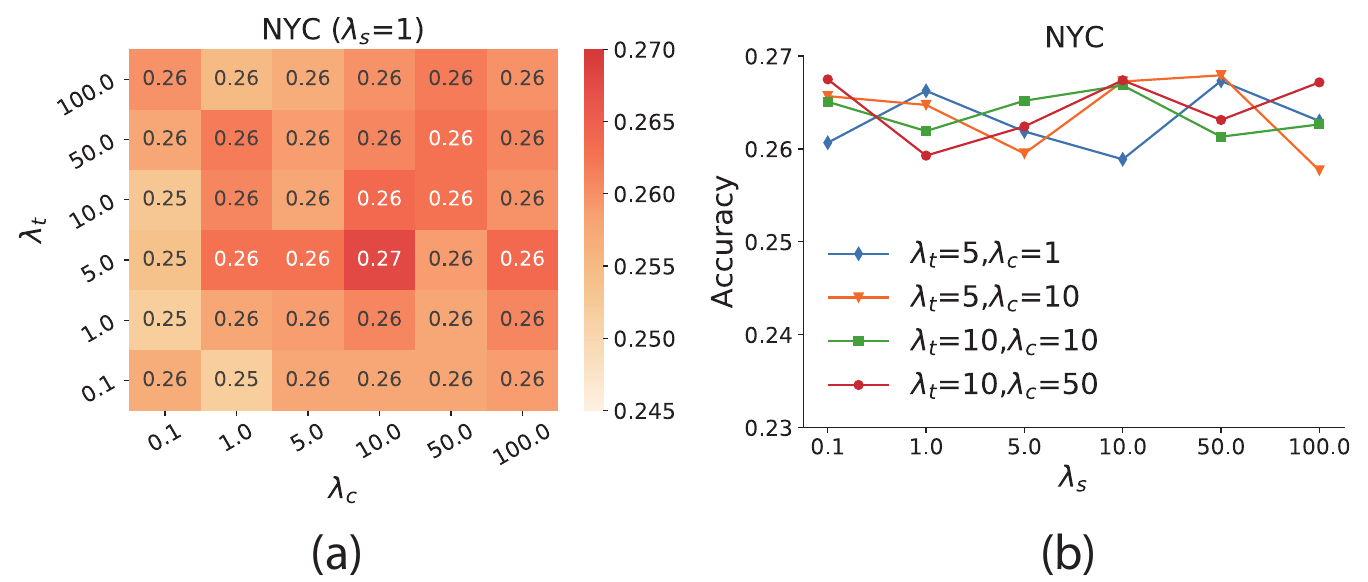}
        \end{center}
        \caption{Analysis of parameter sensitivity.}
        \label{fig:sensitivity}
    \end{figure}

    We perform sensitivity analysis on dataset NYC to examine how the performance of CSLSL is affected by $\lambda_*$, $*\in\{t,c,s\}$. We first vary $\lambda_t$ and $\lambda_c$ to analyze the effect of \textit{time} and \textit{category} prediction subtasks with a fixed $\lambda_s=1$. Then we fix $\lambda_t$ and $\lambda_c$ and vary $\lambda_s$ to observe the impact of spatial-constrained auxiliary tasks. $Recall@1$ is chosen as the evaluation metric and the results of location prediction are averaged of three runs, shown in Figure~\ref{fig:sensitivity}.

    From Figure~\ref{fig:sensitivity} (a), we can observe that the performance of location prediction is more sensitive to $\lambda_c$ than $\lambda_t$, reflecting that the accurate category prediction exerts more influence on the location prediction accuracy, which is also consistent with our proposed decision logic. In addition, the best performance is obtained with $\lambda_t=5$ and $\lambda_c=10$ when $\lambda_s=1$. Figure~\ref{fig:sensitivity} (b) shows that CSLSL reaches a more stable accuracy on NYC when $\lambda_s=5.0$, while the average accuracy is higher when $\lambda_s=10.0$. In summary, CSLSL is robust to these parameters and does not suffer from large performance fluctuations with parameter changes.

\section{Conclusion}
    
    In this work, we propose a Causal and Spatial-constrained Long and Short-Term Learner (CSLSL) to incorporate the individual travel decision logic and the group consistency for next location prediction. In CSLSL, we introduce a causal structure based on multi-task learning to explicitly capture the "\textit{when$\rightarrow$what$\rightarrow$where}" decision logic and enhance location prediction by fully exploiting the temporal and categorized information. We further propose a simple but effective spatial-constrained loss function that acts as a self-supervised auxiliary task to incorporate geographical information and indirectly ensure spatial consistency. We conduct extensive experiments to confirm the effectiveness of the design. 
    In the future, we will assess the scalability of CSLSL in dense mobility datasets with large number of users.
    
\bibliographystyle{ACM-Reference-Format}
\bibliography{reference}
\end{document}